%% file: main.tex
\documentclass[letterpaper, 10pt, conference]{ieeeconf}
\IEEEoverridecommandlockouts
\usepackage{etoolbox} 

\usepackage{amsmath}
\usepackage{amssymb}
\usepackage{float}
\usepackage{graphicx}
\usepackage{subfigure}
\usepackage{svg}
\usepackage[percent]{overpic}
\usepackage[font=small, labelfont=bf, labelsep=period]{caption}
\usepackage{dblfloatfix}

\usepackage{colortbl}
\usepackage{xcolor}
\definecolor{lightgreen}{RGB}{220, 245, 220}
\usepackage{wasysym} 
\usepackage{fontawesome5} 
\usepackage{dirtree}

\usepackage{tabularx}
\usepackage{booktabs}
\usepackage{array}
\usepackage{multirow}
\usepackage{makecell}
\usepackage{threeparttable}
\usepackage{gensymb}

\usepackage[table, xcdraw]{xcolor}
\usepackage{soul, color}

\usepackage{cite}
\makeatletter
\let\NAT@parse\undefined
\makeatother
\usepackage[numbers]{natbib}
\usepackage{svg}
\usepackage{cuted}
\usepackage{caption}

\usepackage{siunitx}
\let\labelindent\relax
\usepackage{enumitem}
\usepackage{lipsum}
\usepackage[colorlinks=true, citecolor=blue, linkcolor=cyan, urlcolor=magenta]{hyperref}

\usepackage{aprl_acronyms}
\usepackage{aprl_misc}

\title{\LARGE \bf Mag4D-SLAM Dataset: A Repeated-Traversal Multi-Modal \\
4D Geomagnetic Dataset for Localization and Mapping}

\author{Bibhutibhusan Nayak$^{1*}$, Hyoseok Ju$^{1*}$, and Giseop Kim$^{1\dagger}$%
\thanks{$^{*}$Equal contribution.}%
\thanks{$^{\dagger}$Corresponding author.}%
\thanks{$^{1}$Bibhutibhusan Nayak, Hyoseok Ju, and Giseop Kim are with the Department of Robotics and Mechatronics Engineering, DGIST, Daegu, Republic of Korea {\tt\small [bibhutibhusan\_223, hyoseokju, gsk]@dgist.ac.kr}.}%
\thanks{This work was supported by the InnoCORE program of the Ministry of Science and ICT (26-InnoCORE-01), by the Institute of Information \& Communications Technology Planning \& Evaluation (IITP) grant funded by the Korea government (MSIT) (No. RS-2025-02219277, AI Star Fellowship Support (DGIST)), and by Basic Science Research Program through the National Research Foundation of Korea (NRF) funded by the Ministry of Education (No. RS-2025-25420118).}%
}

\IEEEaftertitletext{\vspace{-72pt}}

\begin{document}
\maketitle
\vspace{-36mm}   
\input{src/fig/fig_hook}
\thispagestyle{empty} 
\pagestyle{empty}


\maketitle
\thispagestyle{empty}
\pagestyle{empty}

\input{src/abstract.tex}

\input{src/intro}

\input{src/related}
\input{src/sys}
\input{src/calib}

\input{src/data_overview}

\input{src/exp}
\input{src/discussion}
\input{src/conclusion.tex}

\bibliographystyle{unsrt}
{\small
\bibliography{ref}
}

\end{document}

%% file: src/fig/fig_hook.tex
\setlength{\stripsep}{0pt plus 1pt minus 1pt}

\begin{strip}
\centering
\includegraphics[width=\textwidth]{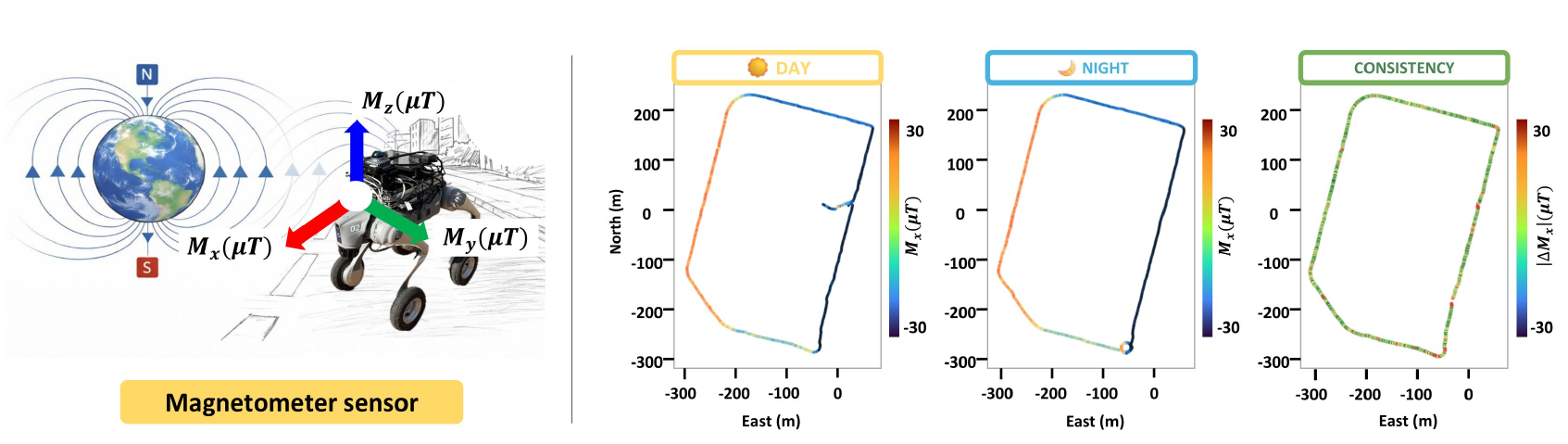}
\captionof{figure}{Overview of the Mag4D-SLAM dataset and sensing paradigm. \textbf{Left:} The Earth's geomagnetic field provides a globally available orientation reference measured by an onboard tri-axis magnetometer ($M_x$, $M_y$, $M_z$) mounted on a quadruped robot platform. \textbf{Right:} Spatial magnetic field maps ($M_x$) along repeated campus trajectories under (left) daytime and (center) nighttime conditions. The rightmost panel shows the cross-session consistency (daytime and nighttime) map ($|\Delta M_x|$), demonstrating that geomagnetic signatures remain spatially stable across day--night sessions, enabling illumination-invariant localization and heading estimation.}
\label{fig:overview}
\vspace{2mm}
\end{strip}

%% file: src/abstract.tex
\begin{abstract}
Geomagnetic sensing offers an infrastructure-free, absolute orientation 
reference that is robust to GNSS denial and visual degradation, yet no 
large-scale outdoor robotics dataset supports its systematic study in SLAM.
Existing magnetic datasets are confined to small-scale indoor environments 
and lack the synchronized multi-modal sensing, repeated-traversal structure, 
and high-precision 6-DoF ground truth required for geomagnetic SLAM research.
We present Mag4D-SLAM, the first large-scale outdoor geomagnetic SLAM dataset. 
It comprises 14 sequences totaling over 18\,km of synchronized LiDAR, camera, IMU,
tri-axis magnetometer, and GNSS measurements with SE(3) ground-truth poses,
collected along structured campus trajectories under paired day/night 
conditions in both forward and reverse directions.
Through repeated-traversal experiments, we analyze three core properties: 
magnetic field repeatability across different recording sessions (daytime and nighttime), drift-free global heading 
estimation, and location-discriminative magnetic signatures for cross-session 
place recognition.
Mag4D-SLAM is designed to support research on yaw drift mitigation, magnetic 
loop closure, and long-term localization --- and to open new research questions 
on how geomagnetic sensing can complement visual and LiDAR modalities or provide
a fallback cue under illumination changes, structural repetition, and GNSS-denied
long-term operation.
\end{abstract}

%% file: src/intro.tex
\section{Introduction}

\ac{SLAM} is a fundamental capability for autonomous robotic systems operating in complex and large-scale environments. Over the past decade, significant advances have been achieved through LiDAR-based \ac{SLAM} \cite{slam-handbook}, visual--inertial odometry, and tightly coupled multi-sensor fusion frameworks. Public benchmarks such as KITTI~\cite{Geiger2013KITTI}, EuRoC~\cite{Burri2016EuRoC}, TUM-VI~\cite{Schubert2018TUMVI}, Oxford RobotCar~\cite{Maddern2017RobotCar}, and Boreas~\cite{Burnett2023Boreas} have played a crucial role in accelerating research by providing synchronized multi-modal data with accurate ground-truth poses. However, these datasets primarily emphasize vision, LiDAR, radar, and inertial sensing, without incorporating geomagnetic measurements as a core modality.

\input{intro_datasettable}

In contrast, the geomagnetic field, despite being globally available,
infrastructure-free, and inherently absolute in orientation, remains
largely absent from robotics-oriented \ac{SLAM} datasets.
Existing magnetic datasets such as UJIIndoorLoc-Mag~\cite{Torres2015UJIIndoorLocMag},
MagPIE~\cite{Hanley2017MagPIE}, and MagWi~\cite{Ashraf2021MagWi} are confined
to indoor environments and lack synchronized visual--LiDAR--IMU--magnetometer measurements
with high-precision 6-DoF ground truth. More recent work such as
Mag-Match~\cite{McDonald2025MagMatch} addresses magnetic map registration but does
not provide a large-scale outdoor dataset for evaluating magnetic-aware \ac{SLAM}.

Beyond datasets, several works have incorporated magnetic information into \ac{SLAM} frameworks. Osman \emph{et al.}~\cite{Osman2022FootSLAM} modeled local magnetic anomalies using Gaussian processes within a Rao--Blackwellized particle filter for pedestrian \ac{SLAM}. Wang \emph{et al.}~\cite{Wang2024AMFSLAM} proposed an orientation-aware 3D \ac{SLAM} system using alternating magnetic fields from powerlines. While these works demonstrate the feasibility of magnetic mapping, they rely on task-specific experimental setups and do not provide standardized robotics datasets that combine camera, IMU, LiDAR, and geomagnetic measurements for reproducible research. \textbf{Across existing magnetic datasets and magnetic \ac{SLAM} studies, three key limitations remain:}

\begin{itemize}
    \item Lack of synchronized conventional robotics sensors (e.g., camera, LiDAR, and IMU) and tri-axis magnetometer measurements,
    \item Absence of structured repeated trajectories for drift and repeatability analysis for the magnetometer measurements in outdoor environments,
    \item No robotics-oriented dataset providing paired multi-modal sensing and accurate ground truth for magnetic-aided \ac{SLAM} research.
\end{itemize}
\noindent Consequently, the role of geomagnetic sensing in mitigating yaw drift, improving loop closure, and supporting multi-modal state estimation remains insufficiently studied in the context of standard \ac{SLAM} datasets.

To address this gap, we introduce the \textbf{Mag4D-\ac{SLAM} Dataset}, a multi-modal geomagnetic dataset designed for magnetic-aware localization and mapping research. The dataset provides synchronized camera, LiDAR, IMU, and tri-axis magnetic measurements recorded along structured circular and rectangular trajectories with accurate ground-truth poses. The ``4D'' designation reflects the spatio-temporal nature of the dataset: the three-axis magnetic field $\mathbf{M}(t) = [M_x(t),\ M_y(t),\ M_z(t)]^\top$ is recorded continuously along repeated traversals across multiple sessions, forming a long-term spatial magnetic field map. Long-term scene understanding and re-localization have been recognized as critical challenges in \ac{SLAM}~\cite{Maddern2017RobotCar, kim2022lt, carlevaris2016university}, yet no existing magnetic dataset provides the multi-session repeated-traversal structure necessary to study geomagnetic sensing as a long-term localization modality.

\input{src/fig/fig_intro_daynight_map}

\figref{fig:overview} illustrates the dataset concept and sensing paradigm. The Earth's geomagnetic field provides a globally consistent yet spatially varying reference, observed in the robot frame as $(M_x, M_y, M_z)$. When recorded along structured trajectories, these measurements form location-dependent magnetic signatures. Despite variations in environmental conditions (e.g., day and night), the underlying magnetic field structure remains largely stable, producing repeatable environmental fingerprints that can help constrain heading drift and improve cross-session consistency in multi-sensor \ac{SLAM}.

\noindent The contributions of this work are summarized as follows:
\begin{itemize}
    \item \textbf{First large-scale outdoor geomagnetic dataset}: Mag4D-\ac{SLAM} provides 14 sequences totaling over 18\,km of synchronized camera, LiDAR, IMU, tri-axis magnetometer, and GNSS data with accurate 6-DoF ground truth, collected on a quadruped robot platform across structured campus trajectories.
    \item \textbf{Repeated-traversal multi-condition design}: The dataset captures paired day and night sessions in both forward and reverse directions along identical paths with mild elevation changes across the trajectory, as shown in \figref{fig:environments}, enabling systematic analysis of geomagnetic consistency, heading-dependent signatures, and illumination-invariant localization.
\item \textbf{Baseline analysis of geomagnetic cues}: We
quantitatively evaluate three core properties relevant to
magnetic-aware localization and mapping: 1) bounded global
heading estimation without accumulated drift, 2) cross-session
magnetic repeatability, and 3) location-discriminative place
recognition. These baselines characterize how geomagnetic
measurements can support future magnetic-aided multi-sensor
fusion and SLAM research.
\end{itemize}

%% file: intro_datasettable.tex
\begin{table*}[t]
\centering
\caption{Comparison of Public SLAM and Magnetic Datasets}
\label{tab:dataset_comparison}
\small
\resizebox{\textwidth}{!}{%
\begin{tabular}{c c c c c | c | c c c c c}
\toprule
\textbf{Year} & \textbf{Dataset} & \textbf{Seq.} & \textbf{Avg. Length} & \textbf{Modalities} & \textbf{Mag.} & \textbf{Environment} & \textbf{Platform} & \textbf{Repeated Trav.} & \textbf{Day/Night} & \textbf{GT Pose} \\
\midrule
2013 & KITTI~\cite{Geiger2013KITTI} & 22 & $\sim$2\,km & L, C, I, G & $\times$ & Outdoor & Car & $\times$ & $\times$ & \cellcolor{lightgreen}$\checkmark$ \\
2016 & EuRoC~\cite{Burri2016EuRoC} & 11 & $\sim$80\,m & C, I & $\times$ & Indoor & MAV & $\times$ & $\times$ & \cellcolor{lightgreen}$\checkmark$ \\
2017 & RobotCar~\cite{Maddern2017RobotCar} & 100+ & $\sim$10\,km & L, C, I, G & $\times$ & Outdoor & Car & \cellcolor{lightgreen}$\checkmark$ & \cellcolor{lightgreen}$\checkmark$ & \cellcolor{lightgreen}$\checkmark$ \\
2017 & MagPIE~\cite{Hanley2017MagPIE} & 10 & $\sim$400\,m & M, I & \cellcolor{lightgreen}$\checkmark$ & Indoor & Handheld & \cellcolor{lightgreen}$\checkmark$ & $\times$ & \cellcolor{lightgreen}$\checkmark$ \\
2018 & TUM VI~\cite{Schubert2018TUMVI} & 28 & $\sim$200\,m & C, I & $\times$ & Indoor/Outdoor & Handheld & $\times$ & $\times$ & \cellcolor{lightgreen}$\checkmark$ \\
2020 & Newer College~\cite{ramezani2020newer} & 4 & $\sim$1.5\,km & L, C, I & $\times$ & Outdoor & Handheld & $\times$ & $\times$ & \cellcolor{lightgreen}$\checkmark$ \\
2021 & MagWi~\cite{Ashraf2021MagWi} & 100+ & Indoor multi-floor & M, W, I & \cellcolor{lightgreen}$\checkmark$ & Indoor & Smartphone & \cellcolor{lightgreen}$\checkmark$ & $\times$ & $\times$ \\
2023 & Boreas~\cite{Burnett2023Boreas} & 40+ & $\sim$8\,km & L, C, R, I, G & $\times$ & Outdoor & Car & \cellcolor{lightgreen}$\checkmark$ & \cellcolor{lightgreen}$\checkmark$ & \cellcolor{lightgreen}$\checkmark$ \\
\midrule
\textbf{2026} & \textbf{Mag4D (Ours)} & \textbf{14} & $\boldsymbol{\sim}$\textbf{1.3\,km} & \textbf{L, C, I, M, G} & \cellcolor{lightgreen}\textbf{$\checkmark$} & \textbf{Outdoor} & \textbf{Quadruped} & \cellcolor{lightgreen}\textbf{$\checkmark$} & \cellcolor{lightgreen}\textbf{$\checkmark$} & \cellcolor{lightgreen}\textbf{$\checkmark$} \\
\bottomrule
\end{tabular}%
}
\begin{tablenotes}
\footnotesize
\item \textit{Abbreviations:}
L = LiDAR, C = Camera, I = IMU, M = Magnetometer,
G = GNSS, R = Radar, W = Wi-Fi.
\end{tablenotes}
\vspace{-3mm}

    \ifdef{\workshopversion}{\vspace{3mm}}{}

\end{table*}

%% file: src/fig/fig_intro_daynight_map.tex
\begin{figure}[t]
    \centering
    \includegraphics[width=\columnwidth]{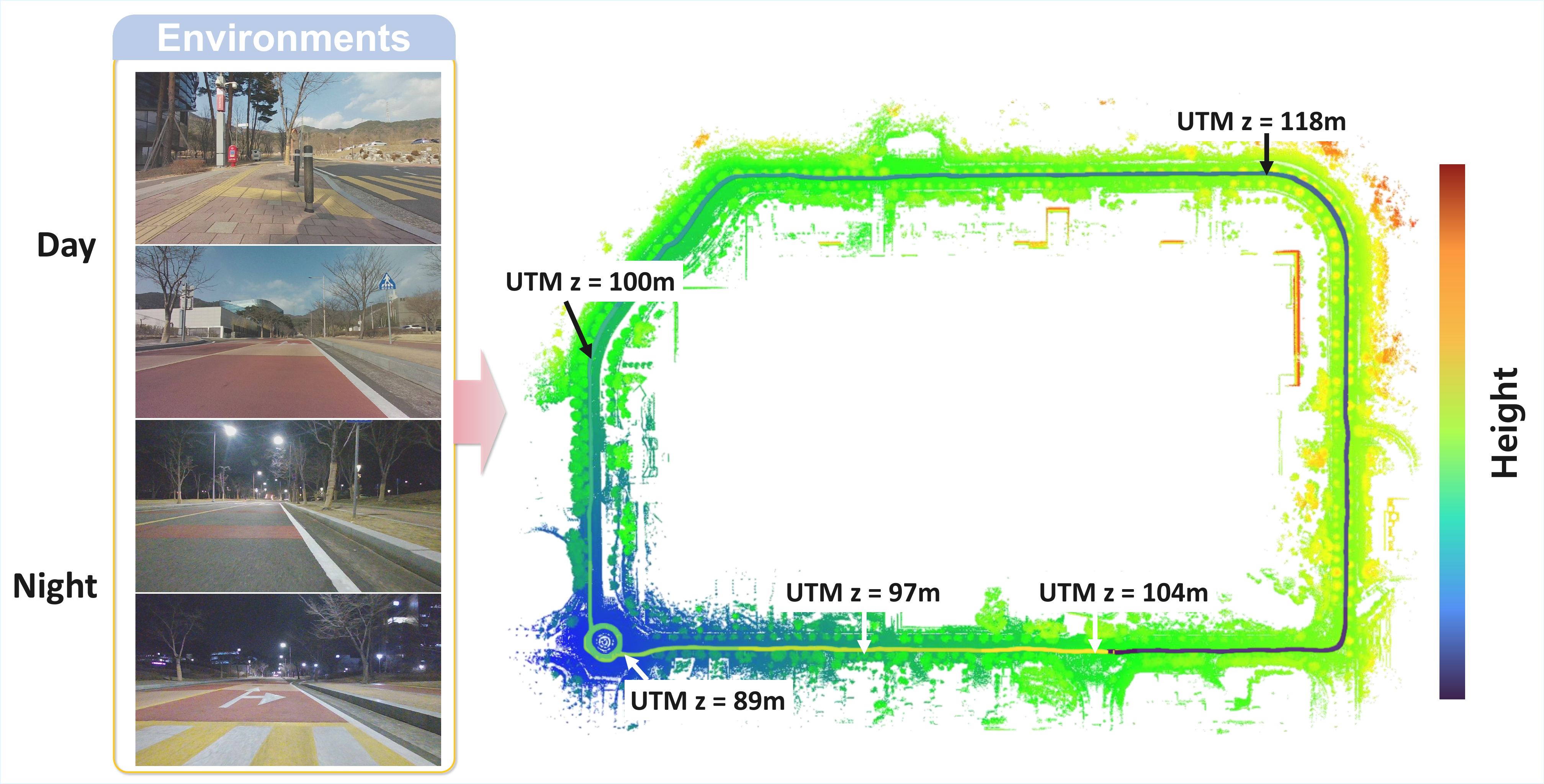}
    \caption{Data collection environments and trajectory overview.
    Left: representative day and night camera images. Right: top-view LiDAR point cloud of the campus loop, colorized by height, showing mild elevation changes across the trajectory.}
    \label{fig:environments}
    \vspace{-5mm}
    \ifdef{\workshopversion}{\vspace{4mm}}{}

\end{figure}

%% file: src/related.tex
\section{Related work}
\subsection{Benchmark Datasets for \ac{SLAM}}
Progress in \ac{SLAM} has been strongly driven by standardized datasets
providing synchronized multi-modal sensing and reliable ground truth.
KITTI~\cite{Geiger2013KITTI} established a large-scale outdoor benchmark
for visual and LiDAR-based SLAM using GPS/INS references.
EuRoC~\cite{Burri2016EuRoC} and TUM-VI~\cite{Schubert2018TUMVI} provide
high-quality visual--inertial data, yet are limited in spatial scale and
primarily indoor scenarios. Oxford RobotCar~\cite{Maddern2017RobotCar}
enables large-scale multi-season evaluation, and Newer
College~\cite{ramezani2020newer} offers centimeter-level ground truth via
map alignment. More recent datasets such as Boreas~\cite{Burnett2023Boreas}
and TartanAir~\cite{wang2020tartanair} further extend diversity across
weather conditions and synthetic environments. Despite their contributions,
none incorporate geomagnetic measurements as a core sensing modality,
limiting their suitability for magnetic-assisted localization and heading estimation.

\subsection{Magnetic Sensing for Localization and \ac{SLAM}}
Magnetic sensing has been explored as a complementary modality where GNSS
is unreliable. Unlike vision or LiDAR, magnetometers capture spatial
distortions of the Earth's magnetic field caused by ferromagnetic
structures, producing location-dependent signatures that support
localization, loop closure detection, and global heading correction.
More recent research has incorporated magnetometers into inertial
navigation systems and probabilistic mapping frameworks for
\ac{SLAM}~\cite{Kok2018MagneticSLAM}, demonstrating that magnetic
anomalies can serve as spatial features for pose correction.
Tan \emph{et al.}~\cite{Tan2020Flydar} proposed a magnetometer-based approach
for high-angular-rate estimation during gyroscope saturation, though their
work addresses dynamic attitude recovery rather than spatial magnetic field
analysis. Magnetometers have also been used to mitigate yaw drift in
visual--inertial pipelines~\cite{joshi2024enhancing}. While these works
demonstrate the utility of magnetic sensing in \ac{SLAM}, they rely on
task-specific setups and lack publicly available multi-modal datasets for
reproducible research.

\subsection{Magnetic Datasets}
Several collections have been proposed to support magnetic-based
localization, yet none fully address the requirements of robotic \ac{SLAM}
research. UJIIndoorLoc-Mag~\cite{Torres2015UJIIndoorLocMag} introduced one
of the first public magnetic fingerprinting databases, containing
magnetometer and IMU measurements along indoor corridors.
MagPIE~\cite{Hanley2017MagPIE} extended this by providing centimeter-level
ground truth across multiple buildings, though it remains focused on
indoor localization without synchronized LiDAR--IMU--magnetometer
measurements. MagWi~\cite{Ashraf2021MagWi} introduced a long-term
magnetic and Wi-Fi dataset across heterogeneous smartphones and
multi-floor buildings, targeting temporal stability rather than robotic
\ac{SLAM} research. More recently, Mag-Match~\cite{McDonald2025MagMatch}
extracts orientation-invariant features from 3D magnetic vector fields via
a physics-informed Gaussian Process for map matching, but does not provide
a large-scale outdoor dataset for magnetic-aware \ac{SLAM}.
As summarized in \tabref{tab:dataset_comparison}, Mag4D-SLAM is designed
to address these gaps by providing large-scale outdoor trajectories,
repeated traversals under day and night conditions, and globally aligned
6-DoF ground-truth poses.

%% file: src/sys.tex
\section{System Overview}
\label{sec:system}

\input{src/fig/fig_sys_platform}
\input{src/tab/tab_sys_sensors}
\input{calib_dataset_quant}
\input{src/fig/fig_calib_calib}
%

\subsection{Sensor Suite}
As shown in \figref{fig:platform}, the Mag4D platform consists of a rigidly mounted multi-modal sensor
payload including a 3D LiDAR, an RGB camera, a 6-axis IMU
(3-axis accelerometer, 3-axis gyroscope), a tri-axis magnetometer,
and a GNSS receiver. Table~\ref{tab:sensors} summarizes the multi-sensor configuration used in our platform with LiDAR--inertial ground-truth pose estimation. All sensors are rigidly attached to a common base
frame and recorded into a single ROS\,2 bag file.

\subsection{Ground Truth Pose Generation}
\label{sec:ground_truth}
Following the approach of the Newer College dataset~\cite{ramezani2020newer},
we generate precise 6-DoF ground-truth poses by registering LiDAR
scans against a high-definition (HD) prior map.
The HD prior map is constructed from mapping-vehicle LiDAR scans using
RTK-GPS fixed-solution trajectory constraints and pose-graph optimization.
Specifically, each LiDAR scan is aligned to the prior map using ICP:
\begin{equation}
    T^* = \arg\min_{T \in SE(3)}
    \sum_{i} \left\| \mathbf{p}_i^{map}
    - T \mathbf{p}_i^{scan} \right\|^2.
\end{equation}
To ensure path smoothness between registered poses,
we used RKO-LIO \cite{malladi2026} as the localizer,
providing continuous LiDAR-inertial estimates
that were subsequently refined via scan-to-map ICP alignment.
Registration accuracy shows a standard deviation below 5\,cm,
consistent with high-precision benchmarks~\cite{ramezani2020newer}.
The ground-truth quality was further verified by overlaying each
trajectory-based session map against the HD prior map in CloudCompare,
with manual inspection confirming spatial alignment
across all sequences.

%% file: src/fig/fig_sys_platform.tex
\begin{figure}[t]
    \centering
    \includegraphics[width=0.7\columnwidth]{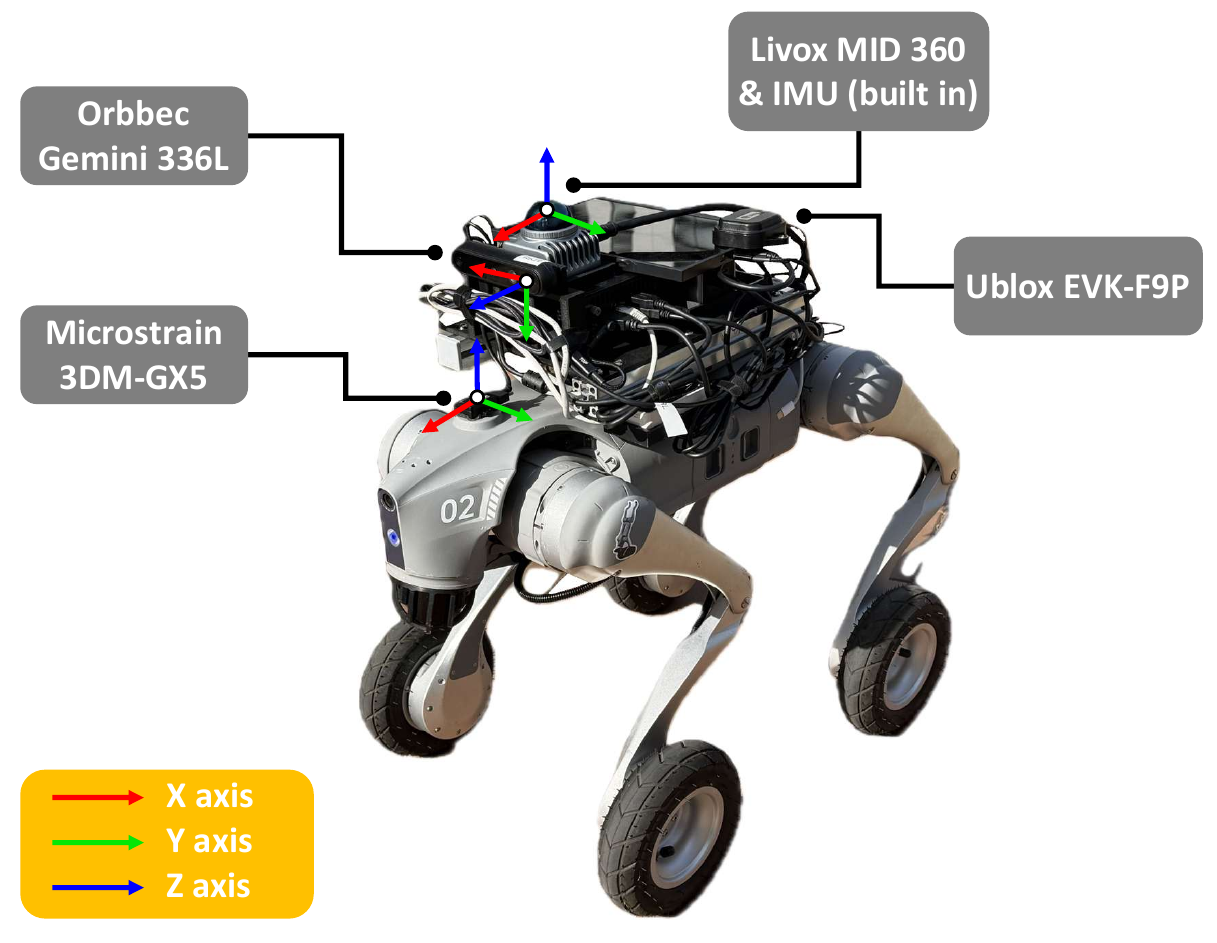}
\caption{Robot platform with all sensors: LiDAR with built-in IMU, GNSS receiver, camera, and magnetometer.}
\label{fig:platform}
\end{figure}

%% file: src/tab/tab_sys_sensors.tex
\begin{table}[t]
\centering
\caption{Overview of the sensor suite used in our system}
\label{tab:sensors}
\vspace{-2mm}
\renewcommand{\arraystretch}{1.1}
\resizebox{\columnwidth}{!}{%
\begin{tabular}{p{1.8cm} p{2.5cm} p{0.8cm} p{4.2cm}}
\toprule
\textbf{Sensor} & \textbf{Type} & \textbf{Rate} & \textbf{Characteristics} \\
\midrule
Camera
  & Gemini 336L
  & 30\,Hz
  & RGB 1280$\times$800,
    H94$^\circ$/V68$^\circ$ \\
\midrule
LiDAR
  & Mid-360
  & 10\,Hz
  & 360$^\circ$/59$^\circ$ FOV;
    200k\,pts/s;
    70\,m range \\
\midrule
IMU
  & Mid-360 (built-in)
  & 200\,Hz
  & 3-axis Accelerometer;
    3-axis Gyroscope \\
\midrule
Magnetometer
  & 3DM-GX5-AHRS
  & 100\,Hz
  &  3-axis,  $\pm$8\,Gauss;   
  Temp.\,compensated \\
\midrule
GNSS
  & EVK-F9P
  & 1\,Hz
  & Multi-band GNSS receiver;
    Operated in standalone mode
    ($\sim$meter-level accuracy) \\
\midrule
SE(3) Ground-truth Pose
  & LiDAR-inertial localization on HD map
  & 10\,Hz
  & Multi-session map-based
    verification (manual) \\
\bottomrule
\end{tabular}
}

\vspace{-2mm}
\end{table}

%% file: calib_dataset_quant.tex
\begin{table*}[t]
\centering
\caption{Sequence-level statistics of our dataset, covering three environments, two lighting conditions, and three traversal directions.}
\label{tab:seq_summary}
\setlength{\tabcolsep}{5pt}
\begin{tabular}{l|c|c|c|r|r}
\toprule
\textbf{Seq.} & \textbf{Env.} & \textbf{Light.} & \textbf{Dir.} & \textbf{Path (m)} & \textbf{Duration (s)} \\
\midrule
\texttt{campus\_day\_fwd\_01}           & Campus        & \faSun\ Day    & $\rightarrow$ Fwd         & 1382.9  & 1203.8 \\
\texttt{campus\_day\_fwd\_04}           & Campus        & \faSun\ Day    & $\rightarrow$ Fwd         & 1554.06 & 1291.9 \\
\texttt{campus\_day\_fwd\_08}           & Campus        & \faSun\ Day    & $\rightarrow$ Fwd         & 1465.62 & 1181.2 \\
\texttt{campus\_day\_fwd\_09}           & Campus        & \faSun\ Day    & $\rightarrow$ Fwd         & 1443.59 & 1132.3 \\
\texttt{campus\_day\_rev\_05}           & Campus        & \faSun\ Day    & $\leftarrow$ Rev          & 1637.24 &  1299.1 \\
\texttt{campus\_day\_rev\_07}           & Campus        & \faSun\ Day    & $\leftarrow$ Rev          & 1605.33 & 1234.9 \\
\midrule
\texttt{campus\_night\_fwd\_00}         & Campus        & \faMoon\ Night & $\rightarrow$ Fwd         & 1376.6  & 1117.3 \\
\texttt{campus\_night\_rev\_11}         & Campus        & \faMoon\ Night & $\leftarrow$ Rev          & 1527.28 & 1224.4 \\
\midrule
\texttt{circle\_day\_fwd\_rev\_02}      & Circle        & \faSun\ Day    & $\leftrightarrow$ Fwd+Rev & 1119.1  &  822.1 \\
\texttt{circle\_day\_fwd\_rev\_06}      & Circle        & \faSun\ Day    & $\leftrightarrow$ Fwd+Rev &  987.53 &  763.0 \\
\texttt{circle\_day\_fwd\_rev\_10}      & Circle        & \faSun\ Day    & $\leftrightarrow$ Fwd+Rev & 1402.79 &  993.2 \\
\texttt{circle\_night\_fwd\_rev\_00}    & Circle        & \faMoon\ Night & $\leftrightarrow$ Fwd+Rev & 1064.4  &  816.9 \\
\midrule
\texttt{library\_day\_fwd\_03}          & Library       & \faSun\ Day    & $\rightarrow$ Fwd         &  189.52 &  163.3 \\
\midrule
\texttt{campus\_circle\_night\_rev\_12} & Campus+Circle & \faMoon\ Night & $\leftarrow$ Rev          & 1676.00 & 1406.2 \\
\bottomrule
\end{tabular}

\vspace{2mm}
\footnotesize
*\texttt{circle\_night\_fwd\_rev\_00} and \texttt{campus\_night\_fwd\_00} were intentionally collected without magnetometer calibration to illustrate the effect of hard-iron and soft-iron distortions, as visualized in \figref{fig:calib}.


\vspace{-5mm}
    \ifdef{\workshopversion}{\vspace{4mm}}{}

\end{table*}

%% file: src/fig/fig_calib_calib.tex
\begin{figure}[t]
    \centering
    \includegraphics[width=1.0\columnwidth]{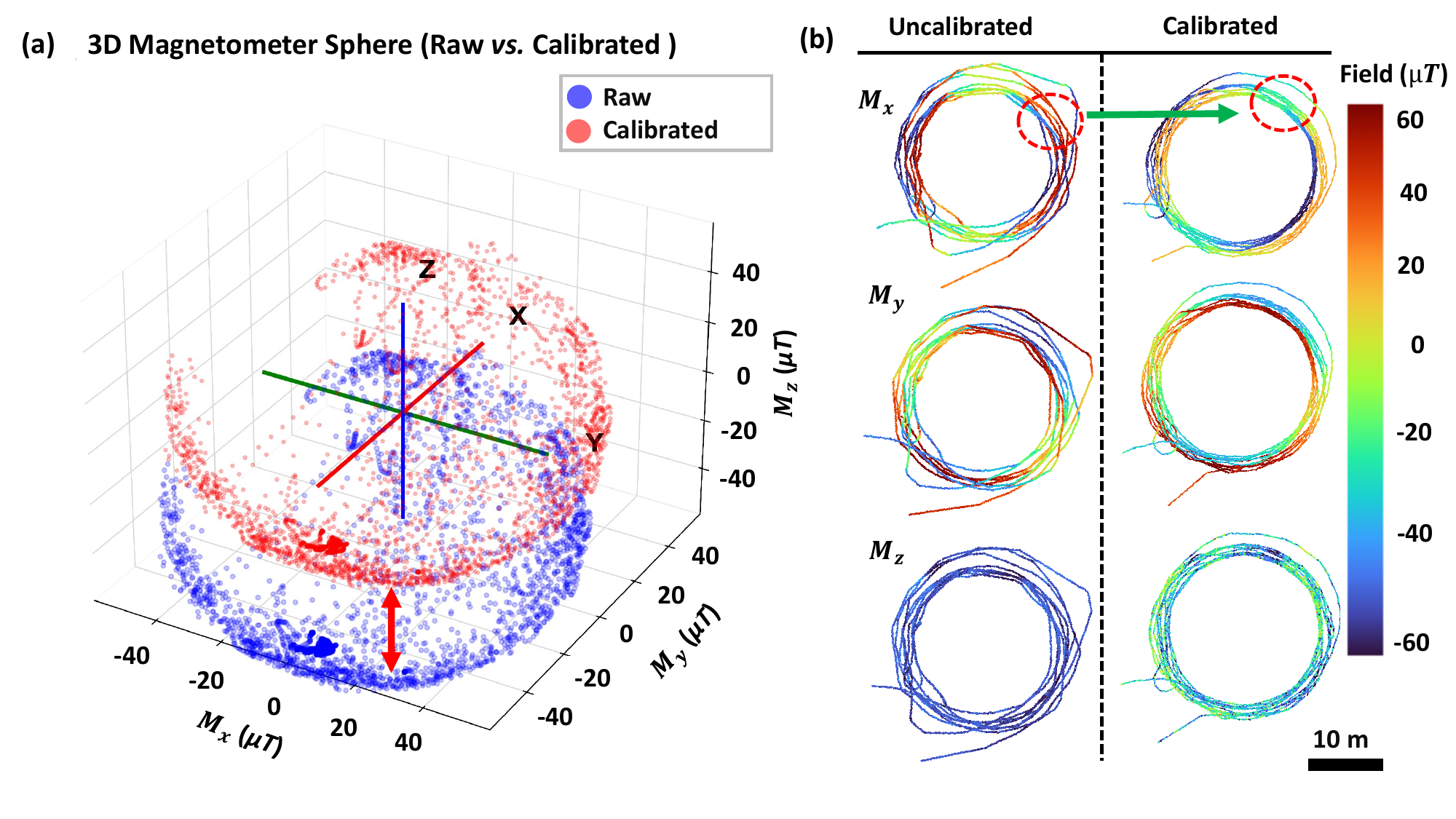}
 
\caption{(a) 3D visualization of uncalibrated (raw) (blue) and calibrated (red) magnetic measurements. 
Calibration removes hard- and soft-iron distortions, transforming the ellipsoidal distribution into a centered sphere. 
(b) 2D projections of $M_x$, $M_y$, and $M_z$, before and after calibration, demonstrating improved symmetry and heading consistency after distortion removal.
}   
    \label{fig:calib}

    \vspace{-3mm}
    \ifdef{\workshopversion}{\vspace{4mm}}{}

\end{figure}

%% file: src/calib.tex
\section{Sensor Calibration}
\label{sec:calibration}

Multi-modal extrinsic calibrations are essential for reliable
sensor fusion and \ac{SLAM} evaluation.
We perform intrinsic, extrinsic, and temporal calibration following established dataset practices~\cite{Schubert2018TUMVI,carlevaris2016university}. Specifically, camera--IMU and LiDAR--IMU extrinsic transformations are estimated
using motion-based hand--eye calibration.
Magnetometer calibration is additionally performed as described below, given its critical role in heading estimation.

\noindent \textbf{Magnetometer calibration background:}
Raw magnetometer readings are affected by hard-iron bias,
soft-iron deformation, axis misalignment, and scale
errors~\cite{Wu2015MagCalibration, joshi2024enhancing}, which introduce systematic
yaw errors over long trajectories without correction.
The measured magnetic vector is modeled as:
\begin{equation}
\mathbf{m}_{meas} = \mathbf{S} \mathbf{m}_{true} + \mathbf{b} + \mathbf{n}
\end{equation}
where $\mathbf{b} \in \mathbb{R}^3$ is the hard-iron bias,
$\mathbf{S} \in \mathbb{R}^{3\times3}$ is the distortion matrix
encoding soft-iron effects, axis misalignment, and scale errors,
and $\mathbf{n}$ is measurement noise.
The calibrated measurement is obtained via ellipsoid fitting~\cite{Vasconcelos2011MagCalibration}:
\begin{equation}
\mathbf{m}_{cal} = \mathbf{S}^{-1}(\mathbf{m}_{meas} - \mathbf{b}),
\qquad \|\mathbf{m}_{\mathrm{cal}}\| \approx B
\end{equation}
where $B$ is the local geomagnetic magnitude.

\noindent \textbf{Magnetometer calibration procedure:}
Calibration was performed in-situ with all onboard electronics
active to capture realistic platform-induced distortions from
motors, batteries, and metallic structures~\cite{Kok2016MagCalibration}.
The robot was manually rotated through full yaw, pitch, and roll
motions to ensure sufficient excitation of all three magnetic
axes for reliable ellipsoid fitting.

\input{src/fig/fig_exp_multi_condition}
\noindent \textbf{Calibration analysis:}
\figref{fig:calib}(a) shows the 3D magnetic measurement
distribution before and after calibration.
The uncalibrated measurements form a distorted ellipsoidal
distribution offset from the origin, indicating hard-iron bias
and soft-iron deformation.
After correction, the measurements approximate a centered sphere,
satisfying $\|\mathbf{m}_{\mathrm{cal}}\|\approx B$. \figref{fig:calib}(b) shows the spatial magnetic
component maps ($M_x$, $M_y$, $M_z$) before and after calibration
along a circular trajectory. The calibrated maps exhibit improved
symmetry and heading consistency, confirming that ellipsoid fitting
effectively compensates for platform-induced distortions.


%% file: src/fig/fig_exp_multi_condition.tex
\begin{figure*}[!t]
    \centering
    \includegraphics[width=0.95 \textwidth]{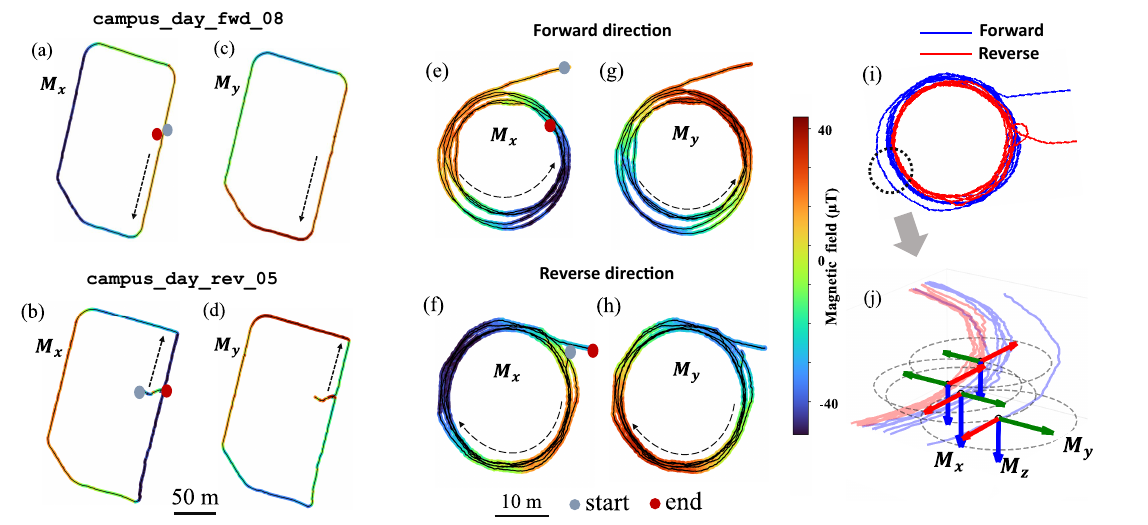}
\caption{
Magnetic field component maps along forward and reverse traversals in the Mag4D-SLAM dataset.
(a--d) Rectangular campus trajectory showing the spatial distribution of $M_x$ and $M_y$ during forward and reverse runs.
(e--h) Circular trajectory executed in forward and reverse directions, illustrating the corresponding magnetic field component maps.
Color represents the magnetic field intensity ($\mu$T), while dashed arrows indicate the direction of motion.
Start and end positions are indicated by gray and red markers.
(i) Trajectory of the forward (blue) and reverse (red) runs near the circle, highlighting traversal direction.
(j) Zoom-in region showing the magnetic field vector at a selected location, where the three-axis components $M_x$, $M_y$, and $M_z$ are visualized using red, green, and blue arrows, respectively.
}
    \label{fig:fig3}
    \vspace{-4mm}
        \ifdef{\workshopversion}{\vspace{4mm}}{}

\end{figure*}

%% file: src/data_overview.tex
\section{Mag4D Dataset}
\label{sec:dataset}

\noindent \textbf{Environment and Trajectory Design: }
Data collection was conducted in a large structured outdoor
environment covering the university campus road segments and pedestrian areas.
The primary trajectory includes a circular road segment
surrounding a central roundabout and rectangular sections
with mild elevation changes, providing realistic terrain
conditions within a controlled geometric layout.
Shorter sequences recorded near the campus library offer
localized high-overlap measurements for fine-grained
spatial consistency analysis.

The dataset includes 14 sequences across repeated traversals
of identical paths under both daytime and nighttime conditions,
with forward and reverse directions as summarized in \tabref{tab:seq_summary}.
While visual sensing performance varies across lighting conditions,
magnetic field measurements remain continuously available,
enabling systematic evaluation of illumination-invariant localization.

\noindent \textbf{Data Format and Availability: }
The Mag4D-SLAM dataset is provided in both raw ROS bag format and
processed CSV format (e.g., \texttt{mag.csv}, \texttt{gt\_pose.csv}). Each sequence contains synchronized
multi-sensor recordings and associated ground-truth files. 
\ifdef{\workshopversion}{}{The dataset is publicly available.\footnote{\url{https://mag4d-dataset.github.io/Mag4d-Dataset/}}}

%% file: src/exp.tex
\section{Experimental Analysis}

This section quantitatively validates three core properties 
of the Mag4D-SLAM dataset: multi-condition magnetic repeatability, 
drift-free heading estimation, and place recognition capability.

\subsection{Multi-Condition Repeatability Analysis}

We evaluate magnetic field variability under forward and reverse traversals for rectangular (campus) trajectories on \texttt{campus\_day\_fwd\_08} and \texttt{campus\_day\_rev\_05}. Fig.~\ref{fig:fig3}(a--d) presents the magnetic field component maps for $M_x$ and $M_y$ along a rectangular campus loop executed in forward and reverse directions. The color scale represents magnetic field intensity ($\mu$T) at each spatial location. Although the geometric trajectory is identical, the measured components depend strongly on the robot's heading, which provides an informative directional signature. This property is particularly valuable when visual SLAM performance degrades under low-light or nighttime conditions.

For the circular trajectory ~\figref{fig:fig3} (e--h), clockwise (reverse) and counter-clockwise (forward) motions from the \texttt{circle\_day\_fwd\_rev\_10} sequence produce distinct magnetic component distributions despite overlapping spatial paths. Specifically, $M_x$ and $M_y$ exhibit sign changes and intensity redistribution along identical spatial segments, with variations spanning approximately $\pm40~\mu$T. These differences arise from changes in body-frame orientation rather than environmental inconsistency. Notably, within a single direction of motion, the magnetic intensity at specific spatial locations remains consistent, demonstrating the uniqueness and repeatability of the magnetic signature.
The zoomed visualization in Fig.~\ref{fig:fig3} (j) further illustrates this behavior: at nearly identical spatial positions, the magnetic vectors (red and green arrows) differ between traversals, with $M_x$ and $M_y$ appearing approximately inverted. This confirms that magnetometer measurements are frame-dependent and strongly influenced by heading, even when the robot revisits the same physical location. Similar orientation-dependent magnetic signatures have also been reported in magnetic field based loop-closure studies~\cite{Kok2024MagSLAM}.

Importantly, despite component-level differences, the overall spatial structure of the magnetic field remains continuous along the trajectory, while heading-dependent differences are particularly pronounced between forward and reverse traversals. The absence of abrupt spatial discontinuities indicates that the environmental magnetic field varies smoothly across the surveyed area. As a result, while the measured components change with robot heading due to projection onto the sensor frame, the underlying magnetic field distribution remains spatially consistent.

These results demonstrate that magnetic measurements are direction-sensitive at the component level yet spatially repeatable in structure, an essential property for magnetic-aware SLAM and heading-consistent loop closure.

\subsection{Quantitative Magnetic Stability Analysis}

\input{src/fig/fig_exp_statistics}

Fig.~\ref{fig:fig5} summarizes the mean and standard deviation of 
$M_x$, $M_y$, $M_z$, and $|M|$ across all sequences.

The first two sequences \texttt{campus\_night\_fwd\_00} and \texttt{circle\_night\_fwd\_rev\_00}, recorded without calibration,
exhibit clear offset distortion. In particular, $M_z$ shows a strong
negative bias of $\approx -50\,\mu$T, while $|M|$ exceeds $60\,\mu$T,
indicating hard-iron and scale effects. The larger spread of $|M|$
confirms violation of the constant-magnitude geomagnetic constraint.

After calibration, the mean components stabilize:
$M_x$ and $M_z$ remain within $\pm15\,\mu$T,
while $M_y$ stays within $\pm10\,\mu$T.
The magnitude $|M|$ consistently clusters between
$35$ and $45\,\mu$T, matching the expected local field strength.

The standard deviation remains consistent across the calibrated runs.
$\sigma_{M_x}$ ranges from $24$--$30\,\mu$T,
$\sigma_{M_y}$ from $18$--$23\,\mu$T,
and $\sigma_{M_z}$ remains lowest at $4$--$7\,\mu$T,
reflecting its vertical alignment with the Earth's field.
The magnitude variation is tightly bounded 
($\sigma_{|M|} \approx 5$--$10\,\mu$T), 
indicating improved spherical consistency.
Overall, calibration enforces $\|\mathbf{m}_{\mathrm{cal}}\| \approx B$
 (local geomagnetic field), 
reduces bias spread, and stabilizes magnetic intensity, 
supporting repeatable signatures for magnetic-aided SLAM.

%
\input{src/exp_heading_obs}
%

\subsection{Place Recognition Baseline}

To assess whether raw magnetic signatures carry location-discriminative 
information, we formulate a null hypothesis $H_0$: retrieval performance 
is equivalent to random selection from the database.
We evaluate cross-session place recognition between 
\texttt{campus\_day\_rev\_05} and \texttt{campus\_night\_rev\_11} 
using nearest-neighbor retrieval with Euclidean distance in magnetic 
feature space. A query pose is considered correctly retrieved if at 
least one of the top-$K$ candidates lies within 3\,m of the ground-truth 
position.
As shown in~\figref{fig:recall}, all three feature configurations substantially exceed the random baseline ($p_0 = K/|\mathcal{D}|$, $|\mathcal{D}|=12{,}245$) across all $K$. At $K=10$, $M_x$+$M_y$+$M_z$ achieves Recall@10\,=\,20.9\% versus a random baseline of 0.08\%, over $250\times$ improvement. A one-sided binomial test yields $p < 10^{-250}$ for all tested $K$, strongly rejecting $H_0$ and confirming that raw magnetic measurements are statistically significant location descriptors. Notably, $M_x$ alone achieves competitive Recall@50 of 47.8\% compared to the full three-axis feature (46.5\%), suggesting that $M_x$ carries the most location-discriminative information among the three components. These results establish a raw-feature baseline and motivate the development of learning-based magnetic descriptors using the Mag4D benchmark.

%% file: src/fig/fig_exp_statistics.tex
\begin{figure}[!t]
    \centering
    \includegraphics[width=\columnwidth]{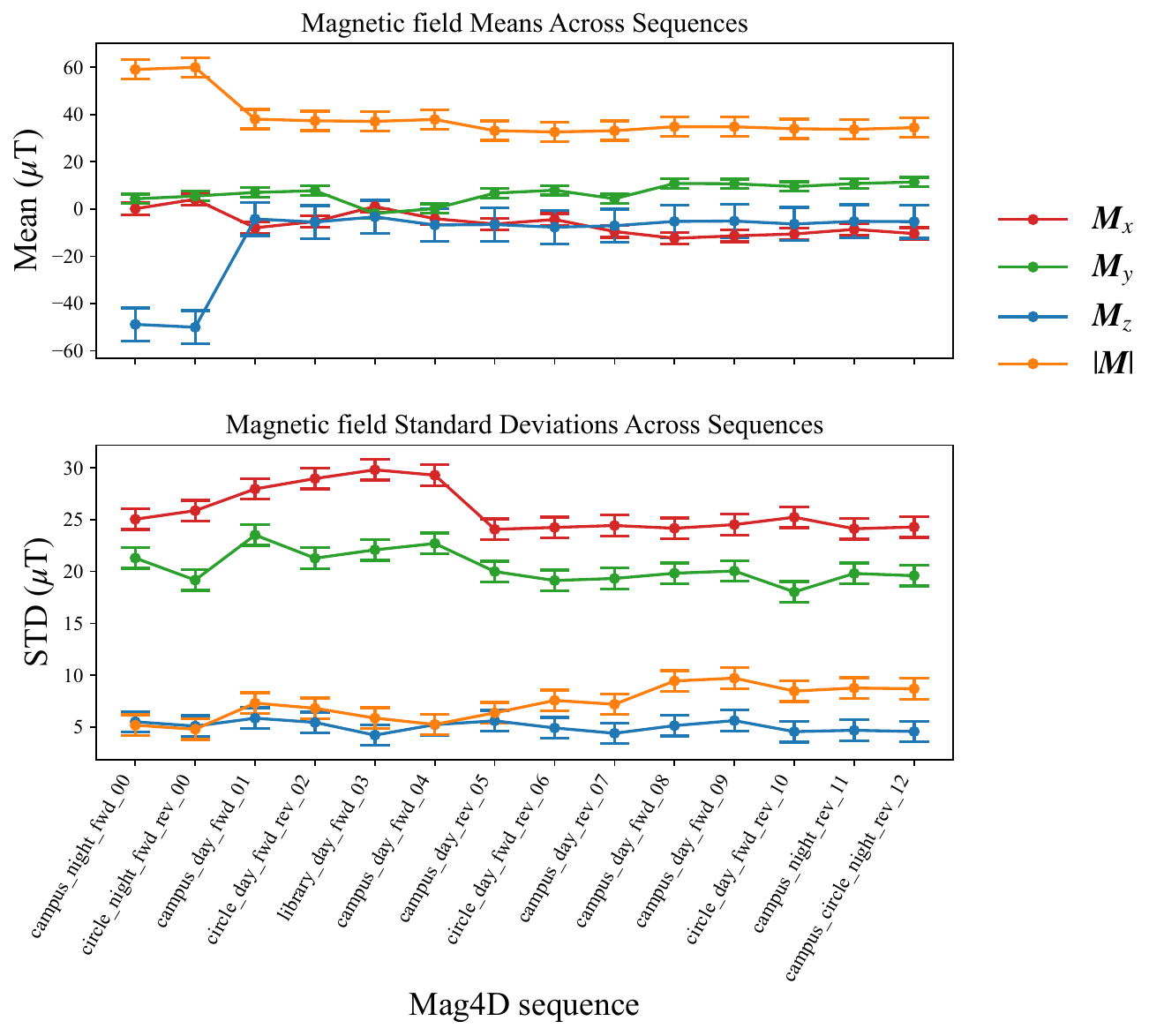}
    \caption{
Magnetic field statistics across all Mag4D-SLAM dataset sequences.
\textbf{Top:} Mean values of the magnetic components ($M_x$, $M_y$, $M_z$) and the magnetic field magnitude ($|M|$) for each sequence, with error bars indicating the standard error. 
\textbf{Bottom:} Standard deviation of the magnetic components and the field magnitude across the same sequences, illustrating the variability of the geomagnetic measurements along different trajectories and environmental conditions (day/night and forward/reverse traversals).
}
\vspace{-6mm}
\label{fig:fig5}    
\end{figure}

%% file: src/exp_heading_obs.tex
\subsection{Drift-Free Heading Estimation}
\label{sec:heading_observability}
A key motivation for Mag4D is to enable global heading
estimation without yaw drift. We quantitatively evaluate
this property by comparing two heading estimation strategies
against the ground-truth (GT) yaw described in \secref{sec:ground_truth}.

\subsubsection{Methods}

\textbf{Direct + Hard-iron Correction.}
We apply tilt-compensated~\cite{caruso1998applications} heading estimation directly from
calibrated magnetic measurements without sequential filtering.
The hard-iron bias is estimated per-sequence using the min/max
method:
\begin{equation}
    b_i = \frac{\max(M_i) + \min(M_i)}{2}, \quad i \in \{x,y,z\}.
    \label{eq:hardiron}
\end{equation}
The tilt-compensated heading is then computed as:
\begin{align}
    M_x^h &= M_x \cos\theta + M_y \sin\phi\sin\theta
             + M_z \cos\phi\sin\theta \notag \\
    M_y^h &= M_y \cos\phi - M_z \sin\phi \notag \\
    \hat{\psi} &= \mathrm{atan2}(-M_y^h,\; M_x^h),
    \label{eq:tilt_comp}
\end{align}
where $\phi$ and $\theta$ denote roll and pitch from the IMU.

\input{src/fig/fig_heading_obs_heading_err}
\input{src/tab/tab_heading_obs_segment_err}
\input{src/tab/tab_heading_obs_heading_err}

\textbf{Direct + Online 2D Circle Fitting.}
Ground robots exhibit small roll/pitch variation
($|\phi| < 5\degree$, $|\theta| < 13\degree$) but full
$360\degree$ yaw coverage during normal traversal, such that
the $M_x$--$M_y$ plane is naturally well-sampled.
We exploit this property to fit a 2D circle online~\cite{renaudin2010complete}:
\begin{align}
    d_k &= \sqrt{(M_x^k - c_x)^2 + (M_y^k - c_y)^2} \notag \\
    (c_x^*, c_y^*, r^*) &= \arg\min_{c_x, c_y, r} \sum_k (d_k - r)^2
    \label{eq:circle_fit}
\end{align}
yielding a calibrated center $(c_x^*, c_y^*)$ subtracted
before applying Eq.~\eqref{eq:tilt_comp}.
Crucially, this requires \emph{no dedicated calibration
maneuver}---normal operation provides sufficient excitation.

\subsubsection{Drift-Free Property}

~\figref{fig:heading_error} illustrates the estimated yaw against GT on \texttt{campus\_circle\_night\_rev\_12}. Table~\ref{tab:heading_results} summarizes RMSE across two sequences. The online 2D circle fitting consistently achieves lower RMSE, with a particularly pronounced improvement on \texttt{campus\_circle\_night\_rev\_12} ($7.7\degree$ vs.\ $12.0\degree$). 

Per-segment RMSE further confirms the drift-free nature of the proposed approach (Table~\ref{tab:segment_rmse}): error remains consistently bounded across all segments regardless of traversal duration.

\subsubsection{Discussion}
The results reveal two key findings. First, online 2D circle-fitting calibration consistently outperforms the standard hard-iron method, particularly on longer sequences, demonstrating that \emph{calibration quality is the dominant factor} in magnetic heading accuracy. Second, ground robots naturally generate sufficient magnetic excitation during normal traversal to enable online calibration, eliminating the need for dedicated calibration maneuvers. The residual error of $\sim\!6\degree$--$8\degree$ is attributable to local magnetic anomalies from campus structures. This analysis confirms that the Mag4D-SLAM dataset provides the structured trajectories and high-precision ground truth necessary to quantitatively evaluate global heading observability---a capability absent from existing \ac{SLAM} benchmarks.

\input{src/fig/fig_heading_obs_recall}
%

%% file: src/fig/fig_heading_obs_heading_err.tex
\begin{figure}[t!]
    \centering
    \includegraphics[width=0.9\columnwidth]{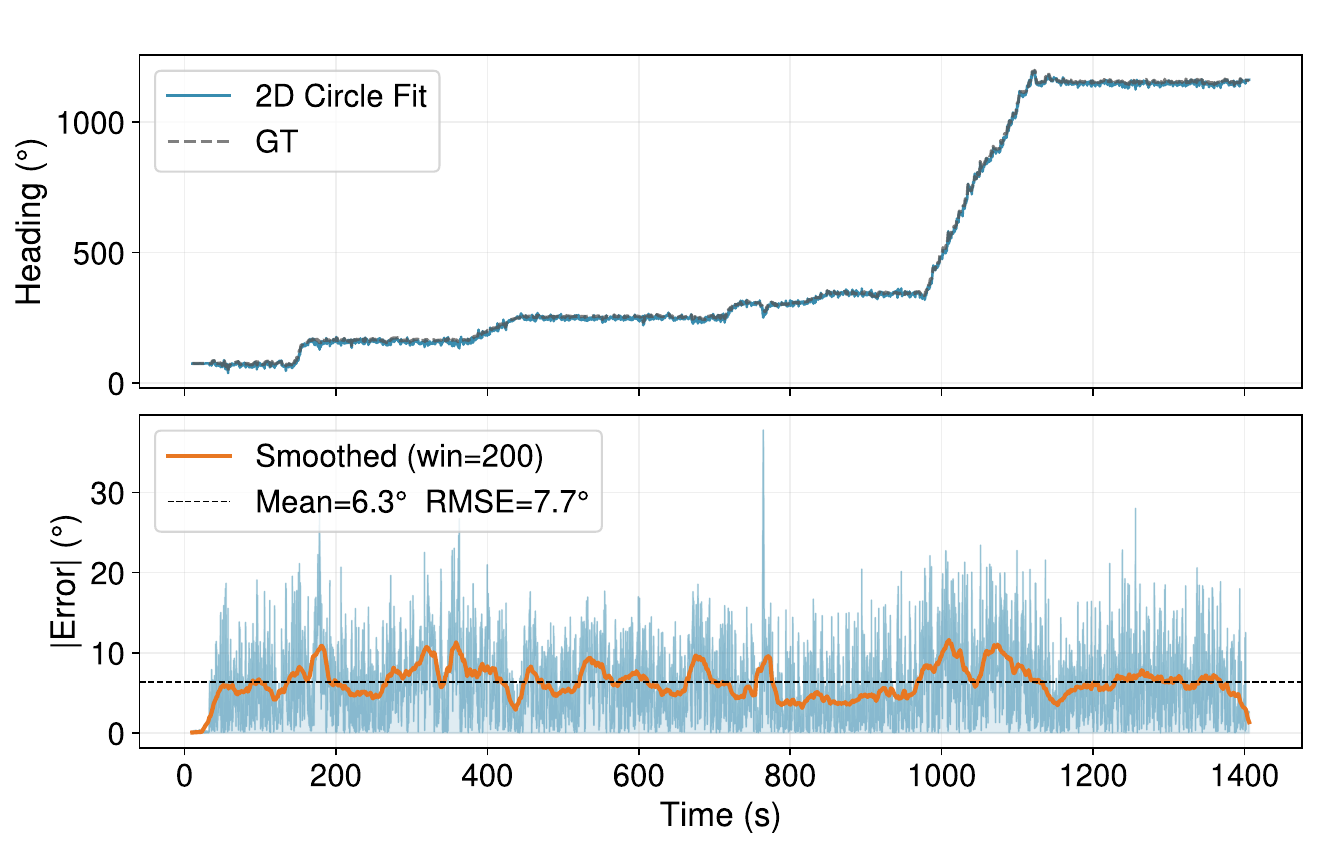}
    \vspace{-5mm}
    \caption{
    Heading estimation on \texttt{campus\_circle\_night\_rev\_12} using the proposed 
    online 2D circle fitting calibration. 
    Top: estimated heading tracks ground truth (GT) continuously 
    without drift over the full 1400\,s trajectory. 
    Bottom: absolute heading error with a moving average filter (window\,=\,200 samples) 
    overlaid; mean error $6.3^\circ$ and RMSE $7.7^\circ$ remain bounded throughout, 
    attributable to local magnetic anomalies along the campus route.
    }
    \label{fig:heading_error}
    \vspace{-1mm}

    \ifdef{\workshopversion}{\vspace{-2mm}}{}

\end{figure}

%% file: src/tab/tab_heading_obs_segment_err.tex
\begin{table}[t!]
\centering
\caption{Per-Segment RMSE Across Representative Sequences}
\label{tab:segment_rmse}
\renewcommand{\arraystretch}{1.18}
\vspace{-1mm}
\setlength{\tabcolsep}{4pt}
\footnotesize
\resizebox{0.95\columnwidth}{!}{%
\begin{tabular}{c|ccc}
\hline
\makecell{\textbf{Sequence} \\ \textbf{Segment}} &
\makecell{\textbf{\texttt{campus\_day}} \\ \textbf{\texttt{\_rev\_05}}} &
\makecell{\textbf{\texttt{campus\_day}} \\ \textbf{\texttt{\_fwd\_08}}} &
\makecell{\textbf{\texttt{campus\_circle\_night}} \\ \textbf{\texttt{\_rev\_12}}} \\
\hline
$0$--$20\%$   & $4.8\degree$ & $4.6\degree$ & $7.1\degree$ \\
$20$--$40\%$  & $5.3\degree$ & $6.2\degree$ & $8.4\degree$ \\
$40$--$60\%$  & $5.6\degree$ & $6.7\degree$ & $7.3\degree$ \\
$60$--$80\%$  & $4.9\degree$ & $7.0\degree$ & $8.3\degree$ \\
$80$--$100\%$ & $5.9\degree$ & $5.7\degree$ & $7.0\degree$ \\
\hline
\end{tabular}%
}
\end{table}

%% file: src/tab/tab_heading_obs_heading_err.tex
\begin{table}[t!]
\centering
\caption{Heading Estimation Accuracy (RMSE)}
\label{tab:heading_results}
\vspace{-2mm}
\renewcommand{\arraystretch}{1.18}
\setlength{\tabcolsep}{4pt}
\footnotesize
\resizebox{0.95\columnwidth}{!}{%
\begin{tabular}{lccc}
\hline
\textbf{Method} & Day (Seq.~05) & Day (Seq.~08) & Night (Seq.~12) \\
\hline
Direct + Hard-iron         & $6.5\degree$ & $6.3\degree$ & $12.0\degree$ \\
\textbf{Direct + 2D Circle} & $\mathbf{5.3\degree}$ & $\mathbf{6.1\degree}$ & $\mathbf{7.7\degree}$ \\
\hline
\end{tabular}%
}
\begin{tablenotes}
\footnotesize
\item Seq.~05: \texttt{campus\_day\_rev\_05}.
\item Seq.~08: \texttt{campus\_day\_fwd\_08}.
\item Seq.~12: \texttt{campus\_circle\_night\_rev\_12}.
\end{tablenotes}
\vspace{-5mm}
\end{table}

%% file: src/fig/fig_heading_obs_recall.tex
\begin{figure}[t!]
    \centering
\includegraphics[width=0.8\columnwidth]{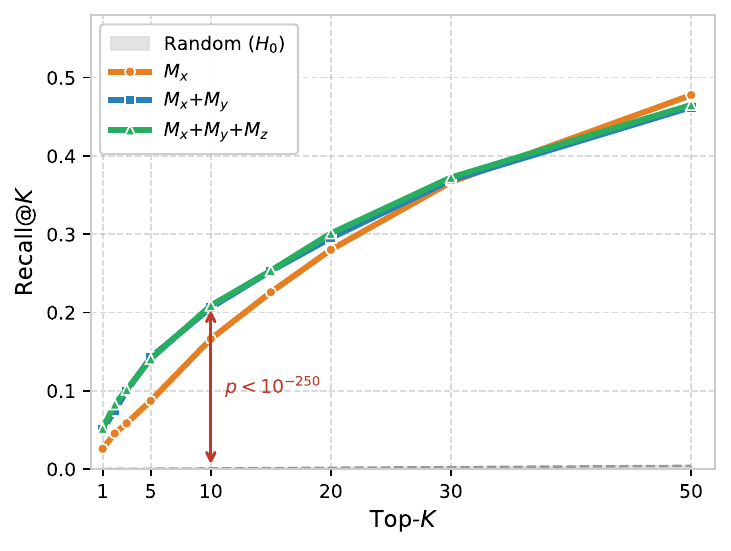}
\vspace{-5mm}
\caption{Recall@$K$ for place recognition using raw magnetic 
features (\texttt{campus\_day\_rev\_05} vs. \texttt{campus\_night\_rev\_11}, day--night cross-session). 
Adding axes progressively improves retrieval performance, 
with $M_x$+$M_y$+$M_z$ achieving Recall@10\,=\,20.9\%, 
demonstrating that raw magnetic signatures carry 
location-discriminative information as a baseline 
for learning-based methods.}    
\vspace{-5mm}
\label{fig:recall}
\end{figure}

%% file: src/discussion.tex
\section{Discussion}
The experimental results demonstrate that geomagnetic sensing provides bounded heading estimates without accumulated drift and location-discriminative signatures across diverse operational conditions. We outline three directions for integrating magnetic sensing into robotic \ac{SLAM} pipelines.

\textbf{Global Heading Correction.} As demonstrated in~\secref{sec:heading_observability}, magnetometer-derived heading achieves RMSE of $6$--$8\degree$ without accumulating drift over trajectories exceeding 1.4\,km, offering a level of long-term consistency that IMU-based orientation cannot provide. This property makes magnetic sensing a practical absolute orientation reference for visual-inertial and LiDAR-inertial odometry systems operating in GNSS-denied environments.

\textbf{Magnetic Place Recognition.} Raw magnetic signatures achieve Recall@10\,=\,20.9\% in cross-session place recognition, providing a viable baseline for magnetic-aided loop closure. In conditions where visual and LiDAR descriptors degrade---including low-light operation, structurally repetitive environments, and adverse weather---magnetic fingerprints offer an infrastructure-free retrieval signal independent of ambient light levels. Learning-based descriptors trained on the Mag4D benchmark hold promise for substantially improving upon this baseline.

\textbf{Long-Term Localization.} Geomagnetic signatures exhibit temporal stability that visual maps often struggle to maintain under visual degradation, seasonal variation, or structural modifications. The cross-session consistency observed in our experiments suggests that a magnetic map constructed from a single traversal can support reliable re-localization across multiple sessions, making magnetic sensing a robust modality for long-term autonomous operation. Nevertheless, environments with strong ferromagnetic interference or dense metallic infrastructure may reduce signature reliability, a limitation that warrants further investigation.

%% file: src/conclusion.tex
\section{Conclusion}
This paper presented the \textbf{Mag4D-SLAM} dataset, a large-scale outdoor magnetic-aware \ac{SLAM} dataset for studying magnetic-assisted localization. Unlike existing magnetic datasets limited to small indoor environments, Mag4D-SLAM provides structured outdoor trajectories with accurate 6-DoF ground truth collected across a university campus under both day and night conditions. The dataset includes synchronized LiDAR, IMU, and tri-axis magnetometer measurements, enabling evaluation of magnetic-assisted heading correction\ifdef{\workshopversion}{}{, place recognition,}
and drift mitigation in modern \ac{SLAM} systems.
Our analysis demonstrates that magnetic signatures remain spatially consistent across repeated traversals and can complement visual--inertial and LiDAR--inertial pipelines for global yaw stabilization. By releasing the Mag4D-SLAM dataset, we aim to support future research on magnetic-aware \ac{SLAM} and encourage the use of geomagnetic sensing as a robust, infrastructure-free modality for long-term robotic autonomy. Although the current dataset includes repeated day/night and directional traversals, it does not yet cover long-term seasonal variations, weather changes, or large temporal gaps across months.
Therefore, the present study primarily focuses on evaluating cross-session magnetic consistency under varying illumination and traversal conditions within a structured outdoor campus environment.
\ifdef{\workshopversion}{Future work includes developing learning-based magnetic descriptors for place recognition and integrating magnetic measurements into multi-sensor \ac{SLAM} pipelines.}{}